\documentclass[10pt,twocolumn,letterpaper]{article}

\usepackage{cvpr}              % To produce the CAMERA-READY version
\usepackage{float}
\usepackage{booktabs}
\usepackage{multirow}
\usepackage[dvipsnames]{xcolor}

\usepackage{setspace}

\definecolor{cvprblue}{rgb}{0.21,0.49,0.74}
\usepackage[pagebackref,breaklinks,colorlinks,citecolor=cvprblue]{hyperref}

\def\paperID{8765} % *** Enter the Paper ID here
\def\confName{CVPR}
\def\confYear{2024}

\title{RecDiffusion: Rectangling for Image Stitching with Diffusion Models}

\author{
    Tianhao Zhou$^{1}$\thanks{Equal contribution.} \hspace{0.3cm}  
    Haipeng Li$^{1}$\footnotemark[1] \hspace{0.3cm} 
    Ziyi Wang$^{1}$ \hspace{0.3cm} 
    Ao Luo$^{2,3}$ \hspace{0.3cm} 
    Chen-Lin Zhang$^{4}$ \hspace{0.3cm} 
    Jiajun Li$^{4}$ \\
    Bing Zeng$^{1}$ \hspace{0.3cm} 
    Shuaicheng Liu$^{1\dagger}$
  \\
    $^{1}$ University of Electronic Science and Technology of China \\
    \quad  $^{2}$ Southwest Jiaotong University \quad $^{3}$ Megvii Technology \quad  $^{4}$ 4Paradigm Inc \\
    {\tt\small \{thzhou,lihaipeng,ziyiwang@std.,eezeng,liushuaicheng@\}uestc.edu.cn,} \\ 
    {\tt\small \{aoluo\}@swjtu.edu.cn, \{zclnjucs,taringlee\}@gmail.com}
}

\begin{document}
\twocolumn[{%
    \maketitle
    \begin{figure}[H]
    \hsize=\textwidth
    \centering
    \includegraphics[width=2.1\linewidth]{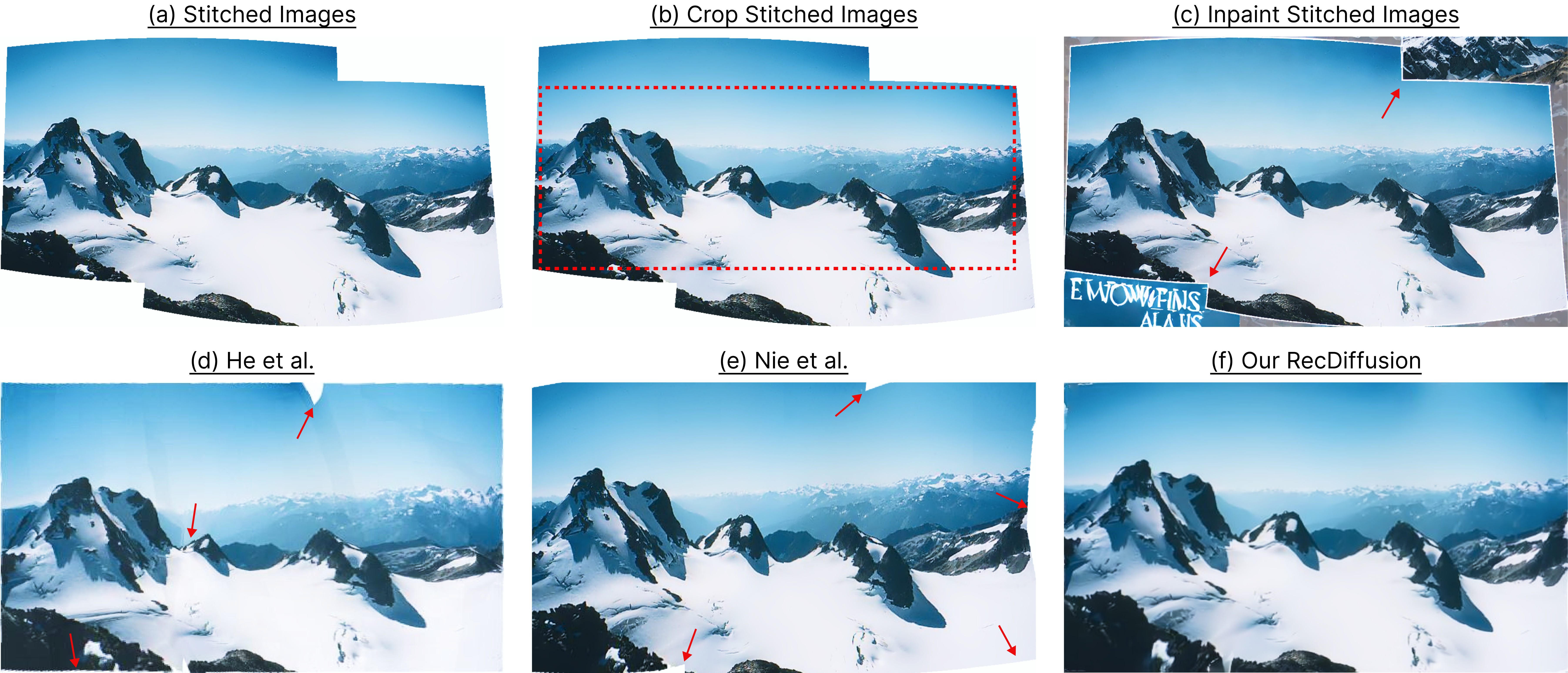}
    \caption{Visual comparisons of our proposed RecDiffusion and previous rectangling approaches including cropping-based, He~\emph{et al.}~\cite{he2013rectangling}, inpainting using Stable Diffusion~\cite{rombach2022high}, and Nie~\emph{et al.}~\cite{nie2022deep}. We can see that simple cropping reduces the field-of-view, the inpainting-based method introduces unsatisfactory extra contents, He~\emph{et al.}~\cite{he2013rectangling} presents distortion and edge artifacts, and Nie~\emph{et al.}~\cite{nie2022deep} unable to maintain a satisfactory rectangular boundary. In contrast, our method properly complements the boundaries and avoids artifacts}
    \label{fig:teaser}
\end{figure}
}]
\renewcommand{\thefootnote}{\fnsymbol{footnote}}
\footnotetext[1]{Equal contribution.}
\footnotetext[2]{Corresponding author.}
\footnotetext[3]{the work is done when Chen-Lin Zhang was in 4paradigm, inc. Chen-Lin Zhang is now at Moonshot AI, Ltd.}

\begin{abstract}
Image stitching from different captures often results in non-rectangular boundaries, which is often considered unappealing. To solve non-rectangular boundaries, current solutions involve cropping, which discards image content, inpainting, which can introduce unrelated content, or warping, which can distort non-linear features and introduce artifacts. To overcome these issues, we introduce a novel diffusion-based learning framework, \textbf{RecDiffusion}, for image stitching rectangling. This framework combines Motion Diffusion Models (MDM) to generate motion fields, effectively transitioning from the stitched image's irregular borders to a geometrically corrected intermediary. Followed by Content Diffusion Models (CDM) for image detail refinement. Notably, our sampling process utilizes a weighted map to identify regions needing correction during each iteration of CDM. Our RecDiffusion ensures geometric accuracy and overall visual appeal, surpassing all previous methods in both quantitative and qualitative measures when evaluated on public benchmarks. Code is released at https://github.com/lhaippp/RecDiffusion.
\end{abstract}

\section{Introduction}
\label{sec:intro}
Image stitching is a technique in which multiple overlapping images of the scene are stitched together to generate an image with a wide field of view (FOV) and high resolution~\cite{szeliski2007image}. These methods often adopt homography~\cite{liu2022content} for global or mesh warps~\cite{zaragoza2013projective} for local alignment of overlapping regions. However, the image boundary produced by stitching algorithms is no longer rectangular due to different capture perspectives, which is unpleasant to view and has been tolerated for quite a long time, as shown in Fig.~\ref{fig:teaser}(a).

To handle such an issue, the most straightforward way is to crop the image by the largest incised rectangle, as shown in Fig.~\ref{fig:teaser}(b). However, the image contents near the boundary have to be discarded, which is also unpleasant due to the loss of image pixels. Another method could be leveraging the recent state-of-the-art (SOTA) generative models to achieve the inpainting, i.e., Stable Diffusion~\cite{rombach2022high}, to complete stitched ones, as illustrated in Fig.~\ref{fig:teaser}(c). However, it introduces extra content which does not belong to the original images. He~\emph{et al.}~\cite{he2013rectangling} proposed the concept of image rectangling, where seam carving technique~\cite{AS07} is adopted for inserting an abundant of seams for an initial rectangular shape, and then a mesh is optimized that warps the image for the final rectangling result. In this way, salient image structures can be preserved, while less important regions are either stretched or squeezed to realize the rectangular shape. However, warping-based methods~\cite{he2013rectangling, he2013content,li2015geodesic} typically preserve only linear structures, such as Manhattan World. Non-linear structures~\cite{zhu2022semi} are usually distorted. An example is shown in Fig.~\ref{fig:teaser}(d).

Recently, Nie~\emph{et al.}~\cite{nie2022deep} proposed a deep learning pipeline that directly minimizes a mesh to warp the image, demonstrating significant improvements over the traditional one~\cite{he2013rectangling}. However, the warping-based method could introduce artifacts and noise due to the lack of accuracy of warping motion fields and the issues inherent in the warping operation~\cite{han2022realflow}, yielding distortions artifacts (inconsistent boundaries and discontinuous lines) as shown in Fig.~\ref{fig:teaser}(e).

In this work, we aim to reformulate the task of image rectangling using diffusion models (DMs)~\cite{sohl2015deep,song2019generative,yang2022diffusion}. Our reasons are twofold: 1) Diffusion models have recently achieved notable performances and demonstrated significant potential in various fields, including, but not limited to, image synthesis~\cite{ho2020denoising,lugmayr2022repaint}, restoration~\cite{wang2022zero,luo2023image,luo2023refusion}, and enhancement~\cite{jiang2023low}. Specifically, DMs have proven to be effective in various motion-related tasks, such as human motion generation~\cite{tevet2022human}, homography synthesis/estimation~\cite{li2024dmhomo}, and depth/optical flow estimation~\cite{saxena2023surprising}; 2) We believe that predicting motion from a single image is an ill-posed problem that can be adequately addressed by DMs. These models have notably improved the outcomes of classical ill-posed problems, like image restoration~\cite{wang2022zero,wang2023unlimited}. Therefore, based on intuition and previous successes, we propose the first diffusion-based learning algorithm as a baseline to tackle the mentioned challenges. Instead of merely seeking a pair of initial and target meshes for warping, we produce the final rectangular results through motion warping operations and image content refinement.

Specifically, the input to the network is the stitched image $I_{\mathbf{S}}$ with irregular boundary, and the output is the corresponding image with rectangular boundary $I_{\mathbf{R}^{\prime}}$. In particular, we first fed $I_{\mathbf{S}}$ into the proposed Motion Diffusion Models (\textbf{MDM}) to produce a motion field. Then we utilize the field to warp $I_{\mathbf{S}}$ to produce a geometrically correct result $I_{\mathbf{\hat{R}}}$, which represents that majority of the content is corrected, but still leaving some details to be optimized, such as the white edges near the boundary, discontinuous lines, and noise. To handle it, we pass $I_{\mathbf{\hat{R}}}$ into another proposed Content Diffusion Model (\textbf{CDM}). To be noticed, the sampling procedure is achieved by fusing $I_{\mathbf{\hat{R}}}$ with the output of CDM. As inspired by Rank-Nullity Theorem~\cite{wang2022zero}, we compute a weighted map $M_{\mathbf{\hat{R}}}$ to identify the confident regions in $I_{\mathbf{\hat{R}}}$, as a result, for every sampling step of CDM, we keep content of $I_{\mathbf{\hat{R}}}$ according to $M_{\mathbf{\hat{R}}}$, and we extract content from CDM's output via $1 - M_{\mathbf{\hat{R}}}$, then they are combined together to be fed into another sampling iteration. With such a strategy, we could generate geometrically accurate and visually pleasing results that outperform all previous methods in quantitative and qualitative comparisons. 

\noindent Our contributions can be summarized as follows:
\begin{itemize}
\vspace{0.5em}
    \item We propose the first diffusion-based framework for image stitching rectangling, namely, \textbf{RecDiffusion}.
    \vspace{0.5em}
    \item  We propose a Motion Diffusion Model (\textbf{MDM}) to generate rectangling motion fields, then a Content Diffusion Model (\textbf{CDM}) to refine image details.
    \vspace{0.5em}
    \item Extensive experiments show that our approach achieves state-of-the-art performance on public benchmarks when compared to previous both traditional and deep methods.
\end{itemize}

\section{Related Work}

\subsection{Image Rectangling}
Image irregular boundary is often produced by applying spatial transformations~\cite{hartley2003multiple}, such as image rotation~\cite{he2013content}, panorama construction~\cite{brown2003recognising}, video stabilization~\cite{Zhang_2023_ICCV} and image stitching~\cite{brown2007automatic}. The most straightforward approach is to crop the empty region for regular boundary. However, some image contents will also be sacrificed. He~\emph{et al}~\cite{he2013rectangling} introduced the image rectangling task, where a mesh is optimized that can warp the image to realize rectangular boundary while retain image contents. Nie~\emph{et al}~\cite{nie2022deep} proposed the first deep learning based rectangling solution, demonstrating superior performances than directly inpaint/synthesize missing regions at borders~\cite{suvorov2022resolution}. Some approaches target at rectangling images under a specific situation, e.g., rotation correction~\cite{he2013content,nie2023deep} and wild-angle rectification \cite{liao2023recrecnet}. Video based methods can further rely on temporal information for compensations~\cite{matsushita2006full,wu2022rectangling}. In this work, we introduce DM for image stitching rectangling.  

\begin{figure*}[t]
\begin{center}
    \includegraphics[width=1\linewidth]{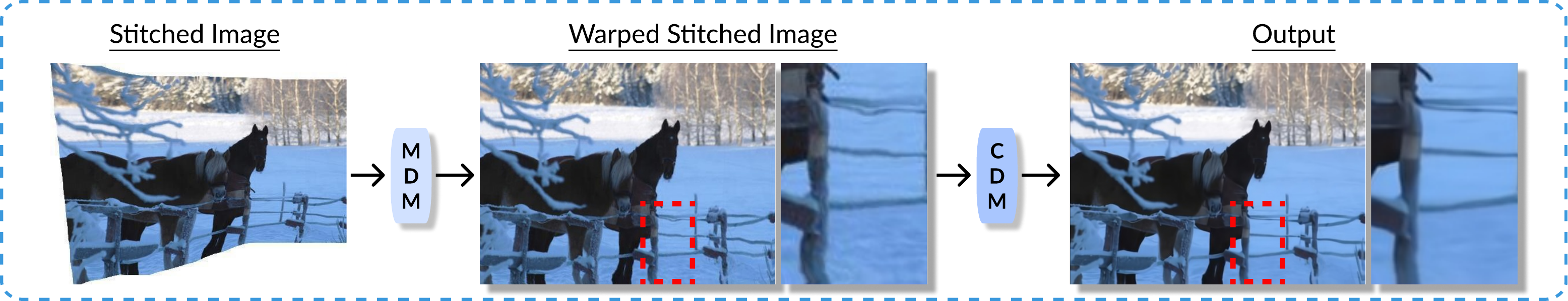}
\end{center}
\vspace{-1em}
\caption{Workflow of RecDiffusion. Initially, Motion Diffusion Models (MDM) are employed to convert irregularly-bordered stitched images into a seamless rectangular form via generated motion fields, which occasionally introduce artifacts like distortion (highlighed by the red box). Content Diffusion Models (CDM) subsequently refine these images.}
\vspace{-1em}
\label{fig:overall_ppl}
\end{figure*}

\subsection{Image Stitching}
Image stitching technique aims to create a larger field of view by combining multiple images of a same scene but captured under different perspectives~\cite{szeliski2007image}. Most of image stitching methods are traditional ones, which concentrate on several different but important aspects, such as effective image feature utilization~\cite{li2015dual,jia2021leveraging}, dealing with large parallax~\cite{zhang2014parallax,lin2016seagull,lee2020warping}, minimizing distortions~\cite{zaragoza2013projective,lin2015adaptive}, preserving shapes of non-overlapping regions~\cite{chang2014shape} and maintaining salient image structures~\cite{lin2016seamless,zhang2020content}. Deep-based methods can improve the performances under challenging conditions, e.g., low or weak textures~\cite{nie2022depth,Nie_2023_ICCV,nie2021unsupervised}. Although many previous works have achieved high quality of stitched images, the shape of the stitching boundary is largely overlooked. In this work, we do not stitch images, but rectanlge the irregular boundary after stitching.    

\subsection{Diffusion Models}
This work is related to Diffusion Models~\cite{yang2022diffusion}, which are generative models, gaining significant popularity recently. DMs work by destroying training data through the successive addition of Gaussian noise, and then learning to recover the data by reversing this noising process~\cite{song2019generative}. DMs can generate data by simply passing randomly sampled noise through the learned denoising process~\cite{liu2023accelerating,ho2020denoising,song2020denoising,wang2022zero}, which iteratively reverse a diffusion process that maps from randomly sampled Gaussian noise to the latent distributions, avoiding issues of instability and model-collapse that often present in previous generative models. Many DMs-based approaches have been proposed with respect to different applications, such as homography synthesis/estimation~\cite{li2024dmhomo}, optical flow estimation~\cite{saxena2023surprising}, human motion synthesis~\cite{tevet2022human}, image restoration/enhancement~\cite{wang2022zero,gao2023implicit,jiang2023low}, 3D model synthesis~\cite{poole2022dreamfusion} and image inpainting~\cite{lugmayr2022repaint,xie2023smartbrush,saharia2022palette,rombach2022high}. In this work, we introduce DMs for the task of rectangling stitched images.

\section{Method}
\label{sec:method}

\subsection{Overview}
\label{sec:overview}
The schema of the processing of stitched images is elucidated in Fig.~\ref{fig:overall_ppl}. Upon the acquisition of stitched images, we process them by leveraging two diffusion models. In its primary stage, Motion Diffusion Models (MDM) generates motion fields that transform stitched images with irregular edges and white margins into seamlessly rectangular formats devoid of these margins, as delineated in Sec.~\ref{sec:RecDiffussor}. The ``image-to-motion" paradigm is adopted during this phase, noted for its efficacy in the delineation of low-level features~\cite{saxena2023surprising}. However, MDM can introduce noise and morphological errors from imperfect motion fields and the complexity of remapping operations~\cite{han2022realflow}, evident in the ``Warped Stitched Image" in Fig.~\ref{fig:overall_ppl}. To ameliorate these artifacts, a secondary phase is invoked, leveraging Content Diffusion Models (CDM), which specifically target the refinement of the images post-MDM application, especially within regions that present issues. This enhancement is done through a novel strategy employing weighted sampling, predicated on the Rank-Nullity Theorem (RNT) principles~\cite{wang2022zero}.

\begin{figure*}[t]
\begin{center}
    \includegraphics[width=\linewidth]{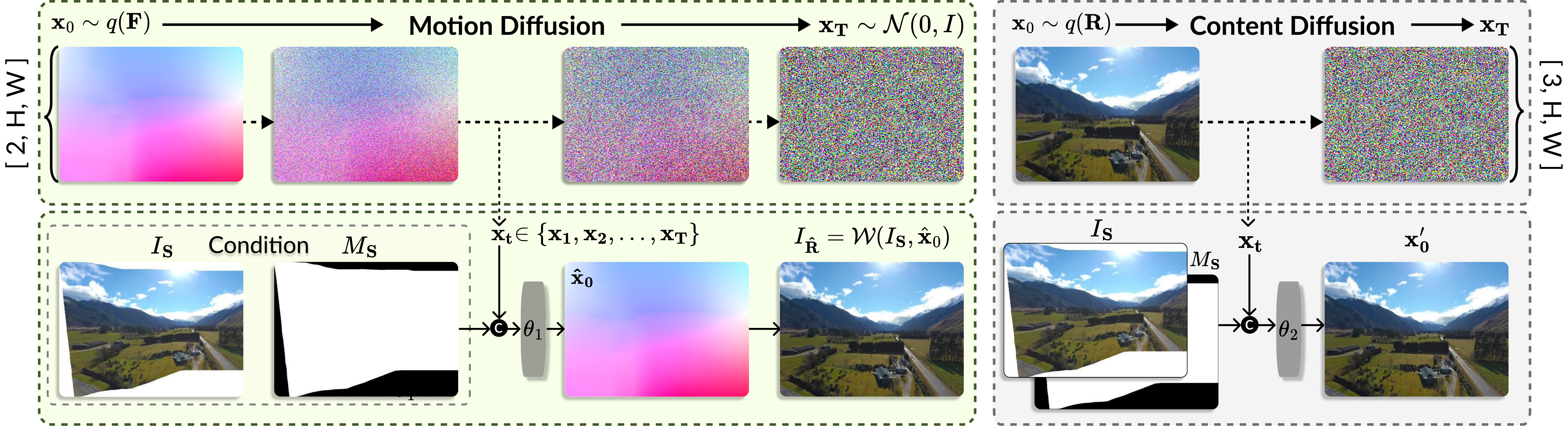}
\end{center}
\vspace{-1em}
\caption{Overview of training procedures. The left block illustrates the training of MDM, which generates motion fields $\mathbf{\hat{x}_0}$ from stitched images $I_{\mathbf{S}}$ and their masks $M_{\mathbf{S}}$, transforming $I_{\mathbf{S}}$ into rectangling images $I_{\mathbf{\hat{R}}}$. The right block shows the training of CDM under the same conditions ($I_{\mathbf{S}}$, $M_{\mathbf{S}}$) to directly generate a rectangling result $\mathbf{{x^{\prime}_0}}$. Both methods aim to reconstruct high-definition rectangling images from stitched inputs, respectively realizing it via motion and content-based manners.}
\vspace{-1em}
\label{fig:main_ppl}
\end{figure*}

\subsection{Diffusion Models}
\label{sec:diffusion_models}

The foundational principles of diffusion models, as explicated by Sohl-Dickstein~\emph{et al.}~\cite{sohl2015deep} and subsequently refined by Ho~\emph{et al.}~\cite{ho2020denoising}, utilize a Markovian transition process over a total of $T$ steps to instigate the sequential infusion of Gaussian noise into an originating data distribution $\mathbf{x}_0 \sim q(\mathbf{x})$. This method generates an array of incrementally noisier images $\{\mathbf{x}_1, \ldots, \mathbf{x}_T\}$, collectively termed forward diffusion, succinctly expressed as follows:

\begin{equation}
q(\mathbf{x}_{1: T} | \mathbf{x}_0) = \prod_{t=1}^T q(\mathbf{x}_t | \mathbf{x}_{t-1}).
\end{equation}

The induction of noise at each interval adheres to a designated Gaussian distribution delineated by a variance schedule $\{\beta_t \in (0,1)\}_{t=1}^T$:
%The induction of noise at each interval adheres to a designated Gaussian distribution delineated by a variance schedule $\{\beta_t \in (0,1)\}_{t=1}^T$:

\begin{equation}
q(\mathbf{x}_t | \mathbf{x}_{t-1}) = \mathcal{N}(\mathbf{x}_t ; \sqrt{1-\beta_t} \mathbf{x}_{t-1}, \beta_t \mathbf{I}).
\end{equation}

Employing the reparameterization technique outlined by Kingma~\emph{et al.}~\cite{kingma2015variational}, it becomes feasible to sample from any intermediary distribution $\mathbf{x}_t$ for an arbitrary $t \in [1,T]$:
\begin{equation}
q(\mathbf{x}_t | \mathbf{x}_0) = \mathcal{N}(\mathbf{x}_t ; \sqrt{\bar{\alpha}_t} \mathbf{x}_0, (1 - \bar{\alpha}_t) \mathbf{I}),
\label{eq:ford_diff}
\end{equation}
where $\alpha_t = 1 - \beta_t$ and $\bar{\alpha}_t = \prod_{i=1}^t \alpha_i$. Thereafter, we introduce an optimized denoising model $\theta$ to inverse the process of diffusion and thereby generate images conforming to a target data distribution, commencing from isotropic Gaussian noise $\mathbf{x}_T \sim \mathcal{N}(\mathbf{0}, \mathbf{I})$:

\begin{equation}
p_\theta(\mathbf{x}_{0:T}) = p(\mathbf{x}_T) \prod_{t=1}^T p_\theta(\mathbf{x}_{t-1} | \mathbf{x}_t),
\end{equation}
\begin{equation}
p_\theta(\mathbf{x}_{t-1} | \mathbf{x}_t) = \mathcal{N}(\mathbf{x}_{t-1} ; \mu_\theta(\mathbf{x}_t, t), \sigma_t^2 \mathbf{I}).
\end{equation}

By executing this inverted transition, the system is endowed with the ability to transform a Gaussian distribution back to the initial data distribution.

To bolster the model's control over the generative procedure and improve the fidelity of the resultant imagery, we introduce additional conditioning variables $\mathbf{y}$ into our architectural framework, following methods advocated by Ho \emph{et al.}~\cite{ho2022classifier}. The conditioning mechanism operates by merging these variables with the intermediary noisy data, yielding enhanced results:

\begin{equation}
p_\theta(\mathbf{x}_{t-1} | \mathbf{x}_t, \mathbf{y})=\mathcal{N}(\mathbf{x}_{t-1} ; \mu_\theta(\mathbf{x}_t, t, \mathbf{y}), \sigma_t^2 \mathbf{I}).
\label{eq:cfg}
\end{equation}

% In our task, we respectively set $\mathbf{x}_0$ to be rectangling images and $\mathbf{y}$ to be stitched images, then we can train a ``image-to-image" diffusion models to achieve the rectangling.

% \begin{equation}
% p_\theta\left(\mathbf{x}_{t-1} \mid \mathbf{x}_t, \mathbf{y}\right) = \mathcal{N}\left(\mathbf{x}_{t-1} ; \boldsymbol{\mu}_\theta(\mathbf{x}_t, t, \mathbf{y}), \

\begin{figure*}[t]
\begin{center}
    \includegraphics[width=1\linewidth]{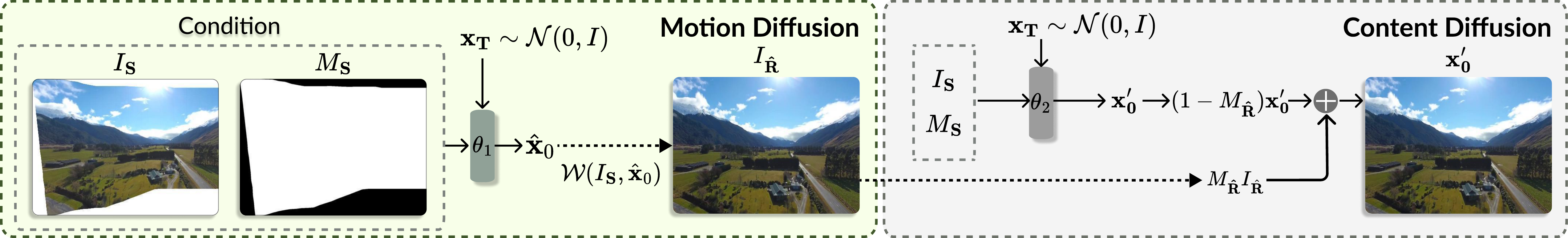}
\end{center}
\vspace{-1em}
\caption{Illustration of the sampling procedure. Initially, stitching images $I_\mathbf{S}$ and masks $M_\mathbf{S}$ are processed by MDM, which generates motion fields $\hat{\mathbf{x}}_0$ iteratively and warps $I_\mathbf{S}$ to form preliminary rectangling images $I_\mathbf{\hat{R}}$ with corresponding confidence masks $M_\mathbf{\hat{R}}$. Secondly, for each sampling step, CDM polishes these images by keeping confidence regions $M_\mathbf{\hat{R}}$ of $I_\mathbf{\hat{R}}$ and updating non-confidence regions $(1-M_\mathbf{\hat{R}})$ via the output of CDM $\mathbf{x}^{\prime}_0$. As a result, we are capable of iteratively reconstructing ideal rectangling images.}
\vspace{-1em}
\label{fig:infe_ppl}
\end{figure*}

\subsection{Rectangling Diffusion Models}
\label{sec:RecDiffussor}

% Based on DDIM~\cite{} and CFG~\cite{}, the framework is capable for rectangling stitched images in a generative way which is different from previous methods. However, the generative capacity of the model is not fully utilized as we generate motion from images, while we do not explicitly constraint the distribution of produced rectangling images. To address the issue, we introduce the Rank-Nullity Theorem (RND).

In our approach, the stitched images denoted as $\mathbf{S}$ can be regarded as a degraded counterpart of rectangling images $\mathbf{R}$, where the composite degradation is attributed to both motion warping and content degradation. Consequently, the proposed framework is designed to learn the transformation from $\mathbf{S}$ back to $\mathbf{R}$, which it achieves by training a motion diffusion model (MDM) and a content diffusion model (CDM) for their respective degradation processes.

\noindent\textbf{Training Process:} We initiate the training of MDM by constructing ``image-to-motion" diffusion models tasked with generating the rectangling motion fields $\mathbf{F}$ that reverse stitched images, $\mathbf{S}$, to rectangling images, $\mathbf{R}$. As depicted in the left block of Fig.~\ref{fig:main_ppl}, based on the conditional framework defined in Eq.~\ref{eq:cfg}, starting from a random sampling of data points, $\mathbf{x}_0 \sim q(\mathbf{F})$, we iteratively introduce noise following Eq.~\ref{eq:ford_diff}. Inputs to the network $\theta_1$ include the associated stitched images $I_{\mathbf{S}}$, their corresponding masks $M_{\mathbf{S}}$ delineating validated image content, along with noised motion fields $\mathbf{x}_\mathbf{t}$. The output of this network, generated motion fields $\mathbf{\hat{x}_0}$, are then utilized to rectangle $I_{\mathbf{S}}$ via a warping function $\mathcal{W}(.)$, thereby yielding the rectangled image $I_{\mathbf{\hat{R}}}$:

\begin{equation}
    I_{\mathbf{\hat{R}}} = \mathcal{W}(I_{\mathbf{S}}, \mathbf{\hat{x}_0}).
\end{equation}

The training loss incorporates two components: the mean square error $\ell_{mse}$ quantifying the divergence between input and output motion fields defined by:
\begin{equation}
    \ell_{mse} = ||\mathbf{\hat{x}_0} - \mathbf{x}_0||^2,
\end{equation}
and a Photometric loss assessing the disparity between resulting rectangled imagery and ground truth, given as:
\begin{equation}
    \ell_{pl} = \left| I_{\mathbf{\hat{R}}} - I_{\mathbf{R}} \right|.
\end{equation}

The composite loss function is therefore presented as a weighted sum:
\begin{equation}
    \ell_{mdm} = \ell_{mse} + \frac{\left|\ell_{mse}\right|}{\left| \ell_{pl} \right|} \cdot \ell_{pl},
\end{equation}
where the norm of $\frac{\left|\ell_{mse}\right|}{\left| \ell_{pl} \right|}$ is used to balance the contribution of each loss component to the overall training objective.

For the training of the Content Diffusion Model (CDM), we employ a parallel strategy, as demonstrated in the right block of Fig.~\ref{fig:main_ppl}. Here, the CDM governs an ``image-to-image" diffusion process involving the model, $\theta_2$, which aims to refine the MDM rectangling images, $I_{\mathbf{\hat{R}}}$. Different from MDM, in CDM, we direct the generation process towards the original rectangling images through sampling, $\mathbf{x}_0 \sim q(\mathbf{R})$, while retaining the same conditional inputs, specifically the stitched images $I_{\mathbf{S}}$ and masks $M_{\mathbf{S}}$. Consequently, the model produces an enhanced version of the rectangling images, denoted as $\mathbf{{x^{\prime}_0}}$. The associated training loss is the MSELoss measuring the distance between these enhanced images and the ground truth rectangling images:

\begin{equation}
    \ell_{cdm} = || \mathbf{{x^{\prime}_0}} - \mathbf{x}_0 ||^2.
\end{equation}
In harnessing power of diffusion models to capture and correct motion-based and content-based degradations, the integrated training process enables the reconstruction of high-fidelity rectangling images from input stitched counterparts.

\noindent\textbf{Sampling Process:} Fig.~\ref{fig:infe_ppl} delineates the procedure we follow for the sampling process. After adequately training both the Motion Diffusion Model (MDM) and the Content Diffusion Model (CDM), we progress through two principal steps to transform stitching images towards refined rectangling images. Initially, pairs of stitching images $I_{\mathbf{S}}$ along with their corresponding masks $M_{\mathbf{S}}$ serve as the input conditions. From here, the correction motion fields $\hat{\mathbf{x}}_0$ are iteratively estimated from Gaussian noise over a series of steps. These fields then warp $I_{\mathbf{S}}$ to produce preliminary rectangling results, $I_{\mathbf{\hat{R}}}$. Considering that this process may introduce noise and artifacts due to the accuracy of the generated motion fields and the properties of the remapping operation, our strategy includes computing a confidence mask, $M_{\mathbf{\hat{R}}}$, to categorize regions according to their reliability in terms of confidence levels.

More specifically, $M_{\mathbf{\hat{R}}}$ is computed via 3 different masks: 1) input stitched image masks $M_{\mathbf{S}}$, 2) intensity map of $\hat{\mathbf{x}}_0$, $M_0$, which is obtained by the normalized displacement of grids, 3) white edge mask of $I_{\mathbf{\hat{R}}}$ as $M_1$. Then we can formulate $M_{\mathbf{\hat{R}}}$ as:
\begin{equation}
    \label{eq:cdm_mask}
        M_{\mathbf{\hat{R}}} = 1 - \max\left\{M_1, \frac {\omega_0 \cdot 1 + M_0 + M_{\mathbf{S}}} {\omega_0 + 2}\right\},
\end{equation}
where $\omega_0$ is a hyper-parameter to be tuned.

Secondly, we utilize CDM to refine the noise and artifacts present within $I_{\mathbf{\hat{R}}}$. To facilitate this, we deploy a weighted sampling technique inspired by the Rank-Nullity Theorem (RNT)~\cite{wang2022zero}. We commence by establishing two primary constraints: a consistency constraint (Eq.~\ref{eq:consistency}), which asserts that $\mathbf{\hat{r}}$ (representative of vectorized $I_{\mathbf{\hat{R}}}$) should match the vectorized desired rectangling images $\mathbf{{r^\prime}}$, after the degradation $\mathbf{A}$. Moreover, we implement a realism constraint (Eq.~\ref{eq: realness}), proposing that the generated results for $\mathbf{{r^\prime}}$ conform with the expected distribution:

\begin{equation}
    Consistency: \mathbf{\hat{r}} = \mathbf{A} \mathbf{{r^\prime}}, 
    \label{eq:consistency}
\end{equation}
\begin{equation}
    Realism: \mathbf{{r^\prime}} \sim q \left( \mathbf{R} \right).
    \label{eq: realness}
\end{equation}

We then formulate an equation based on Rank-Nullity considerations (Eq.~\ref{eq:rnd}) that serves to merge the constraints of consistency and realism:
\begin{equation}
    \mathbf{{r^\prime}} = \mathbf{A}^{\dagger} \mathbf{A} \mathbf{{r^\prime}} + \left( \mathbf{I} - \mathbf{A}^{\dagger} \mathbf{A} \right) \mathbf{{r^\prime}},
    \label{eq:rnd}
\end{equation}
where $\mathbf{A}^{\dagger}$ represents the Pseudo-inverse of $\mathbf{A}$. This equation expresses $\mathbf{r'}$ as a combination of its projection into the range-space of a matrix $\mathbf{A}$ and its projection into the corresponding null-space.

Back to our method, we desire to produce favorable rectangling images $I_{\mathbf{R}'}$ (where $\mathbf{r'}$ represents the vectorizing format) via RNT. To achieve it, we consider the confident regions $M_{\mathbf{\hat{R}}}$ of $I_{\mathbf{\hat{R}}}$ to be the range space, and the rest regions as the null space. Therefore, the degradation matrix $\mathbf{A}$ is replaced with confidence masks, producing a novel relationship between $\mathbf{\hat{r}}$ and $\mathbf{r'}$ for each sample step as Eq.~\ref{eq:simpli_rnd}, in which the multiplication between $\mathbf{M}$ and $\mathbf{r}$ is element-wise: 
\begin{equation}
    \label{eq:simpli_rnd}
    \mathbf{{r^{\prime}}} = \mathbf{M}\sqrt{\bar{\alpha}_{t}}\mathbf{\hat{r}}  + \left( \mathbf{I} - {\mathbf{M}} \right) \mathbf{{r^\prime}},
\end{equation}
in essence, the diagonal matrix $\mathbf{M}$ that arises from vectorizing $M_{\mathbf{\hat{R}}}$ helps integrate the confidence levels associated with different regions of the image. With this revised relationship, the refinement process iteratively adjusts $\mathbf{\hat{r}}$ towards the ultimate target, $\mathbf{r'}$.

This iterative process is graphically depicted in the right panel of Fig.~\ref{fig:infe_ppl}. Specifically, for each iteration, the algorithm preserves pixels of $\mathbf{\hat{r}}$ with high confidence, as indicated by the mask $\mathbf{M}$, which is equal to $M_{\mathbf{\hat{R}}}I_{\mathbf{\hat{R}}}$, and is then diffused to timestep $t$ by multiplying $\sqrt{\bar{\alpha}_t}$. Conversely, for the remaining pixels identified by $(\mathbf{I} - \mathbf{M})$, the output of the CDM is used to substitute values with the goal of reducing noise and enhancing realism. It can be realized as $(1 - M_{\mathbf{\hat{R}}})\mathbf{x}_0^\prime$. Such a sampling approach allows us to progressively reconstruct $\mathbf{r'}$, thus achieving step-by-step refinement of $\mathbf{x}_0^\prime$, accomplishing the rectangling images $I_{\mathbf{R}'}$.

\section{Experiment}
\label{sec:experiment}

We provide configurations of the experiment in Sec.~\ref{sec:implement_details}. The comparisons with other methods are shown in Sec.~\ref{sec:quanti_comp} and Sec.~\ref{sec:quali_comp}. In addition, we test the generalizability of the models in Sec.~\ref{sec:generaliza_exp} and compare with inpainting methods in Sec.~\ref{sec:impaint_comp}. Lastly we conduct ablation studies in Sec.~\ref{sec:abla_study}. Furthermore, we provide a brief introduction to datasets and dynamic visualizations in \textbf{Supplementary Materials} to better demonstrate the results.

\subsection{Implementation Details}
\label{sec:implement_details}
The proposed framework consists of MDM and CDM. The design of them follows DDIM~\cite{song2020denoising} and classifier-free method (CFG)~\cite{ho2022classifier}. Both of them are trained using Adam optimizer~\cite{kingma2017adam} with parameters $\beta_{1}=0.9$, $\beta_{2}=0.99$. The generated pseudo motion fields used to train MDM is from the previous state-of-the-art method, i.e., Nie~\emph{et al.}~\cite{nie2022deep}. For the configuration of MDM, the condition scaling of CFG is 6, learning rate is $2.0 \times 10^{-4}$, batch size is 64, sampling step is 2, the number of training steps is $320,000$. For CDM, the batch size is 32, learning rate is $1.0 \times 10^{-5}$, sampling step is 200, and the number of training steps is $450,000$. The time taken to train on 8 NVIDIA A100s are 3 and 4 days for MDM and CDM, respectively. More details will be demonstrated in Supplementary Materials.

\begin{figure*}[t]
\begin{center}
    \includegraphics[width=\linewidth]{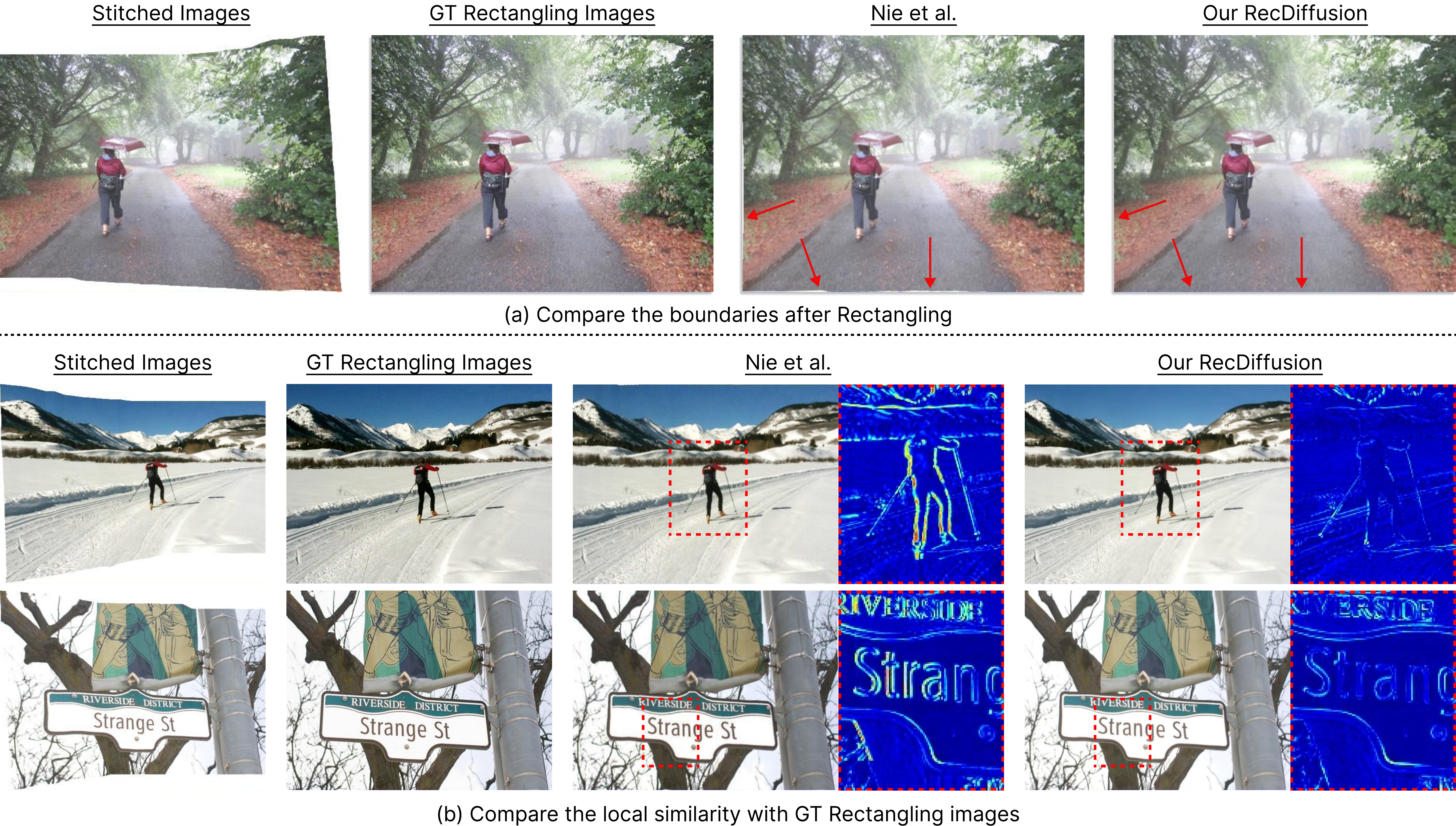}
\end{center}
\vspace{-1em}
\caption{Comparative Evaluation of Nie~\emph{et al.}~\cite{nie2022deep} on the DIR-D Dataset. The input stitched images and the GT rectangling references are displayed in the first two columns. The third column shows the rectangling results by Nie~\emph{et al.}, while our proposed diffusion models-based outcomes are exhibited in the last column. In figure (a), red arrows accentuate white edge artifacts present in the outcomes of the previous state-of-the-art. Figure (b) scrutinizes the presence of internal artifacts such as line discontinuities and local distortions, highlighted within Regions of Interest (ROIs) circled on alignment heatmaps where darker shades signal higher fidelity to the ground truth. Our results demonstrate enhanced similarity to the ground truth, indicating a significant reduction in artifacts compared to the previous method.}
\label{fig:quali_white_edges}
\end{figure*}

\subsection{Quantitative Comparison}\label{sec:quanti_comp}
\begin{table}[!t]
    \centering
    \def\temptablewidth{0.45\textwidth}
    {\rule{\temptablewidth}{1pt}}  %根据使用情况灵活设置，线的粗细
    \begin{tabular*}{\temptablewidth}{@{\extracolsep{\fill}}l|ccc}
    \hline
    Method & FID $\downarrow$ & SSIM $\uparrow$ & PSNR $\uparrow$ \\ 
    \hline
    Reference & 12.25 & 0.3245 & 11.30\\
    \hline
    He~\emph{et al.}~\cite{he2013rectangling}& {-} & {0.3775} & {14.70}\\
    \hline
    Nie~\emph{et al.}~\cite{nie2022deep}& {4.14} & {0.7141} & {21.28}\\
    \hline
    Ours& \textbf{3.63} & \textbf{0.7733} & \textbf{22.21}\\
    \hline
    \end{tabular*}
    {\rule{\temptablewidth}{1pt}}
    \caption{Quantitative comparisons of PSNR, SSIM, and FID between our method and other rectangling methods on the DIR-D~\cite{nie2022deep} test set. ``Reference" denotes that the metrics are computed by using input stitched images as rectangling results. The best results are highlighted in \textbf{bold}.}
    \vspace{-1.5em}
    \label{table:quantity_comp} 
\end{table}

We adopt the evaluation settings from previous studies, utilizing the Fréchet inception distance (FID), Structural Similarity Index (SSIM), and Peak Signal-to-Noise Ratio (PSNR) to assess these methods. Our evaluation on the \textbf{DIR-D} dataset, presented in Table \ref{table:quantity_comp}, compares our approach with both the traditional rectangling method~\cite{he2013rectangling} and deep learning-based technique~\cite{nie2022deep}. Specifically, we calculate FID on trainset as 519 test cases are not enough to compute a meaningful score. Previous methods tend to treat rectangling as a regression problem, addressing it with specialized architectures and task-specific loss functions, such as local-to-global strategies, feature warps, perception loss, or grid constraints. In contrast, our generative framework does not rely on specialized components nor regression frameworks, relying exclusively on diffusion models and achieving superior performance across all metrics, establishing a new state-of-the-art. It offers a novel potential technological path to solving the problem.

\begin{figure*}[t]
\begin{center}
    \includegraphics[width=1\linewidth]{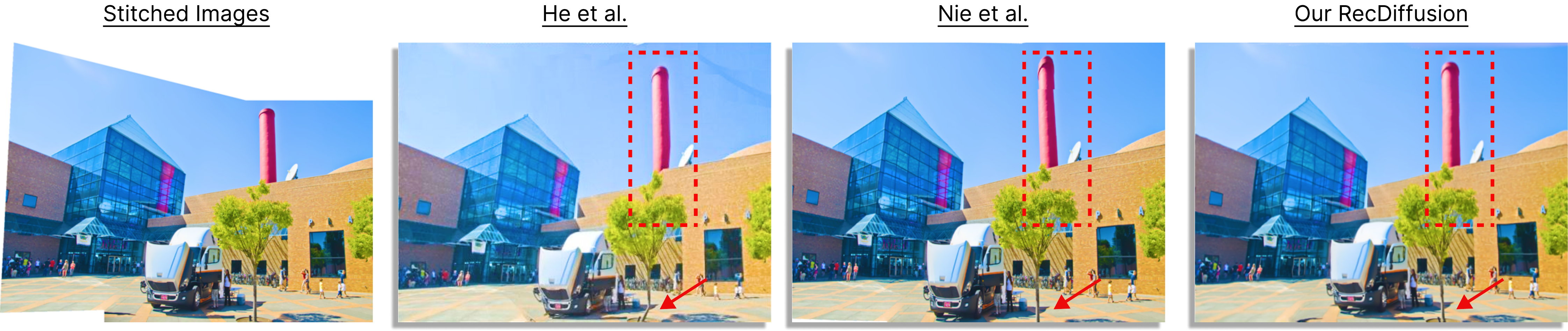}
\end{center}
% \vspace{-1em}
\caption{We test the zero-shot capacity of different methods on \textbf{APAP-conssite}~\cite{zaragoza2013projective}, including He~\emph{et al.}~\cite{he2013rectangling}, Nie~\emph{et al.}~\cite{nie2022deep} and our RecDiffusion, trained on \textbf{DIR-D}~\cite{nie2022deep}. Roofs and branches twisted (red boxes and arrows in He~\emph{et al.} result), chimney breakage and flower bed moved out of figure (red boxes and arrows in Nie~\emph{et al.} result) exist in their outputs. Our results performs the best among them.}
\vspace{-1em}
\label{fig:generalize}
\end{figure*}

\begin{figure}[t]
\begin{center}
    \includegraphics[width=1\linewidth]{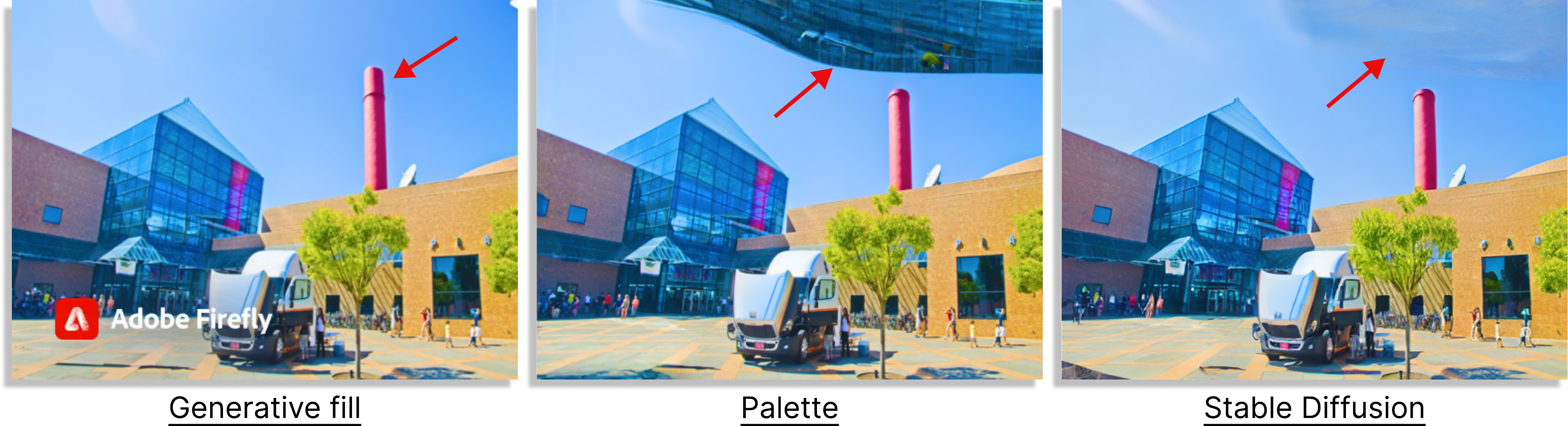}
\end{center}
% \vspace{-1em}
\caption{Inpainted stitching images by Adobe commercial software - Generative fill, Palette~\cite{saharia2022palette} and Stable Diffusion 2.1~\cite{rombach2022high}.}
% \vspace{-1em}
\label{fig:inpaint}
\end{figure}

\begin{figure}[t]
\begin{center}
    \includegraphics[width=1\linewidth]{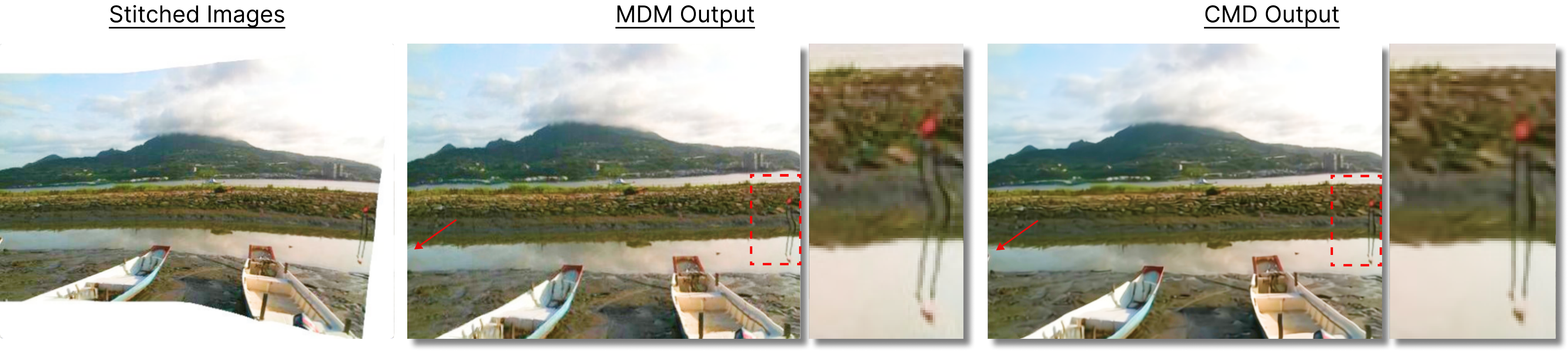}
\end{center}
% \vspace{-1em}
\caption{Illustration of the output of MDM and CDM. We can find that the local distortion is well handled by CDM.}
\label{fig:ablat_mdm_cdm}
\end{figure}

\subsection{Qualitative Comparison}
\label{sec:quali_comp}
Our method is evaluated against the previous state-of-the-art method on \textbf{DIR-D}~\cite{nie2022deep}. The visual comparisons are respectively illustrated in Fig.~\ref{fig:quali_white_edges}. For comparisons in Fig.~\ref{fig:quali_white_edges} (a), we mainly compare whether the corrected stitched images are seamless rectangular ones or not, because as far as we know one of the most key aspects of the rectangling task is the complete elimination of irregular boundaries of the stitched images. However, Nie~\emph{et al.}~\cite{nie2022deep} leveraging the warping meshes to achieve rectangling, naturally faces the risk of irregular boundary artifacts due to the accuracy of correcting motion and the inherent problems with warp operations. We use red arrows to indicate those white edging regions in the figure. On the contrary, our diffusion models-based framework locates the issue at the schematic side and is capable of generating desired rectangling images.

On the other hand, despite the incomplete white edges, artifacts could occur within the images. For example, line discontinuities and local distortions can occur, due to the lack of accuracy and smoothness of the warping motion fields. We demonstrate related images in the Fig.~\ref{fig:quali_white_edges} (b). More specifically, to vividly demonstrate the similarities between produced results and GT images, we adopt the alignment heatmap~\cite{jiang2023semi}, where darker regions correspond to better similarity. We encircle some of the ROIs in the graphs, which are subject contents. From the results, we can observe that our produced results are closer to the GT rectangling images, thus suffering from less artifacts. More dynamic results in GIF format can be found in \textbf{Supplementary Materials}.

\subsection{Generalizability Experiments}
\label{sec:generaliza_exp}

Experiments involve zero-shot inference on \textbf{APAP-conssite} dataset~\cite{zaragoza2013projective} using He~\emph{et al.}~\cite{he2013rectangling}, while Nie~\emph{et al.}~\cite{nie2022deep} and our RecDiffusion are pre-trained on the \textbf{DIR-D} dataset~\cite{nie2022deep}. Outcomes are illustrated in Fig.~\ref{fig:generalize}. From the results, we observe that other methods produce artifacts as highlighted by red box and arrow, for example, chimneys and branches are twisted in the result of He~\emph{et al.}. The output of Nie~\emph{et al.} also contains line discontinuity (red box) and flower bed is removed from the bottom of the figure (red arrow). On the contrary, our framework's robust backbone ensures its generalizability across different datasets.

\subsection{Comparison with Inpainting Methods}
\label{sec:impaint_comp}

\begin{table}[!h]
    \centering
    \def\temptablewidth{0.45\textwidth}
    {\rule{\temptablewidth}{1pt}}  %根据使用情况灵活设置，线的粗细
    \begin{tabular*}{\temptablewidth}{@{\extracolsep{\fill}}l|ccc}
    \hline
    Method & FID $\downarrow$ & SSIM $\uparrow$ & PSNR $\uparrow$ \\ 
    \hline
    Reference & 12.25 & 0.3245 & 11.30\\
    \hline
    Palette~\cite{saharia2022palette}& {-} & {0.3315} & {14.49}\\
    \hline
    Stable Diffusion 2.1~\cite{rombach2022high} & 15.58 & {0.3276} & {14.23}\\
    \hline
    Ours& 3.63 & 0.7733 & 22.21\\
    \hline
    \end{tabular*}
    {\rule{\temptablewidth}{1pt}}
    \caption{Quantitative comparisons of PSNR, SSIM, and FID with inpainting methods on the DIR-D dataset.}
    \vspace{-1em}
    \label{table:inpaint_quanti} 
\end{table}

Image rectangling aims to eliminate irregular boundaries while maintaining as much data consistency and achieving good qualitative results as possible, therefore, previous methods~\cite{he2013rectangling, nie2022deep} choose to warp stitching images. While inpainting methods~\cite{rombach2022high,saharia2022palette} are powerful at generating visually pleasing outcomes, they tend to introduce extra content into the stitched ones, as demonstrated in Fig.~\ref{fig:inpaint}, thus affecting the data consistency negatively. As shown in Table~\ref{table:inpaint_quanti}, inpainted stitching images (row 2 and 3) result in the much lower PSNR/SSIM metrics than by RecDiffusion. Moreover, their FID scores (computed on the trainset) are higher than the FID scores comparing the stitched input images to the ground truth rectangling images, indicating a significant discrepancy in image quality.

\subsection{Ablation Study}
\label{sec:abla_study}
We evaluate our framework designs through experiments on test set of \textbf{DIR-D} dataset~\cite{nie2022deep}, starting with comparisons under the details of Motion Diffusion Models (MDM). Specifically, we conduct experiments on different resolutions and the effectiveness of stitched image masks $M_{\mathbf{{S}}}$ as conditions. Then we explore the design of Content Diffusion Models (MDM), we evaluate different combinations, including solely leveraging CDM, streamlining MDM with CDM, and the effectiveness of weight sampling masks.

%\vspace{-0.5em}

\subsubsection{Motion Diffusion Models}
\label{sec:abla_mdm}

While implementing MDM, conditional stitched image masks $M_{\mathbf{{S}}}$ and resolution are important factors impacting performance as shown in Table~\ref{table:ablt_mdm}. Without the mask, the model cannot even outperform the baseline, i.e., the model to generate pseudo-motion fields for the train set. The larger resolution delivers better results as well as expected, but smaller resolution could lead to much faster inference.

\begin{table}[h]
\centering
\def\temptablewidth{0.45\textwidth}
{\rule{\temptablewidth}{1pt}}  %根据使用情况灵活设置，线的粗细

\begin{tabular*}{\temptablewidth}{@{\extracolsep{\fill}}cc|cc}
\hline
Condition Mask &Resolution &SSIM$\uparrow$& PSNR$\uparrow$\\ 
\hline
  &$256\times192$ & {0.6125} & {20.23}\\
\hline
 $\checkmark$ &$256\times192$ & {0.7337} & {21.97}\\
\hline
$\checkmark$ & $512\times384$ & \textbf{0.7580} & \textbf{22.03}\\
\hline
\end{tabular*}

{\rule{\temptablewidth}{1pt}}
\caption{Comparison of different resolutions and conditions.}
\vspace{-0.5em}
\label{table:ablt_mdm} 
%\vspace{-0.6em}
\end{table}

% \vspace{-2em}

\subsubsection{Content Diffusion Models}
\label{sec:abla_cdm}
Results are demonstrated in Table~\ref{table:ablat_cdm}. We find that CDM alone is ineffective as it generates images with different illuminations. Besides, we find that the output of MDM could be directly improved via CDM, and the weighted sampling mask (WSM) could further improves the performance as illustrated in Fig.~\ref{fig:ablat_mdm_cdm}, where local distortions are eliminated, and missing content has been restored (mark by red arrow).

\begin{table}[!h]
\centering
\def\temptablewidth{0.45\textwidth}
{\rule{\temptablewidth}{1pt}}  %根据使用情况灵活设置，线的粗细

\begin{tabular*}{\temptablewidth}{@{\extracolsep{\fill}}ccc|cc}
\hline
MDM & CDM & WSM & SSIM $\uparrow$ & PSNR $\uparrow$ \\ 
% \hline
% $\checkmark$& & & {0.7580} & {22.03}\\
\hline
& $\checkmark$& & {0.3129} & {14.70}\\
\hline
$\checkmark$& $\checkmark$& & {0.7618} & {22.03}\\
\hline
$\checkmark$& $\checkmark$& $\checkmark$& \textbf{0.7733} & \textbf{22.21}\\
\hline
\end{tabular*}

{\rule{\temptablewidth}{1pt}}
\caption{Comparison of the effectiveness of CDM and WSM.}
% \vspace{-1em}
\label{table:ablat_cdm} 
%\vspace{-0.6em}
\end{table}

\vspace{-1em}
\section{Conclusion}

In this work, we present \textbf{RecDiffusion}, the first diffusion models-based approach for rectangling stitched images. Compared to previous methods specialized for this task, which include special network structures and loss functions, we demonstrate that a typical diffusion model based on generative motion outperforms these methods. Furthermore, to address the problem of artifacts introduced by motion inaccuracy and the warping operation, we propose a strategy that uses a weighted sampling mask. This strategy combines the advantages of warping methods and generative modeling, effectively improving performance. This approach could potentially be applied to other motion-related tasks. Overall, we have achieved state-of-the-art performance in comparison to previous methods on public benchmarks. Code and model weights are available at https://github.com/lhaippp/RecDiffusion.
\newline
\\
\noindent\textbf{Acknowledgements.} This work was supported by National Natural Science Foundation of China (NSFC) under grant No.62372091, 61872405, the ``111” Project under Grant B17008 and Sichuan Science and Technology Program of China under grant No.2023NSFSC0462.
\clearpage

{
    \small
    \bibliographystyle{ieeenat_fullname}
    \bibliography{main}
}


\end{document}